%% file: main.tex
\documentclass[runningheads]{llncs}
\usepackage{graphicx}

\usepackage{tikz}
\usepackage{comment}
\usepackage{amsmath,amssymb} 
\usepackage{color}

\usepackage[accsupp]{axessibility}  

\usepackage[caption=false]{subfig}
\usepackage{bm}
\usepackage{booktabs}
\usepackage{multirow}
\usepackage{multibib}
\newcites{supp}{Supplementary References}

\newcommand{\R}{\mathbb{R}}

\begin{document}
\pagestyle{headings}
\mainmatter
\def\ECCVSubNumber{7709}  

\title{On the benefits of knowledge distillation for adversarial robustness} 

\titlerunning{On the benefits of knowledge distillation for adversarial robustness}
%
\author{Javier Maroto
\and Guillermo Ortiz-Jiménez
\and Pascal Frossard
}
\authorrunning{J. Maroto et al.}
%
\institute{EPFL, Switzerland}
\maketitle

\begin{abstract}

Knowledge distillation is normally used to compress a big network, or teacher, onto a smaller one, the student, by training it to match its outputs. Recently, some works have shown that robustness against adversarial attacks can also be distilled effectively to achieve good rates of robustness on mobile-friendly models. In this work, however, we take a different point of view, and show that knowledge distillation can be used directly to boost the performance of state-of-the-art models in adversarial robustness. In this sense, we present a thorough analysis and provide general guidelines to distill knowledge from a robust teacher and boost the clean and adversarial performance of a student model even further. To that end, we present Adversarial Knowledge Distillation (AKD), a new framework to improve a model's robust performance, consisting on adversarially training a student on a mixture of the original labels and the teacher outputs. Through carefully controlled ablation studies, we show that using early-stopping, model ensembles and weak adversarial training are key techniques to maximize performance of the student, and show that these insights generalize across different robust distillation techniques. Finally, we provide insights on the effect of robust knowledge distillation on the dynamics of the student network, and show that AKD mostly improves the calibration of the network and modify its training dynamics on samples that the model finds difficult to learn, or even memorize.

\end{abstract}

\section{Introduction}
\label{sec:intro}

Knowledge distillation (KD)~\cite{hinton2015distilling} is a common technique used to compress neural networks to be deployed on computationally constrained devices~\cite{vanhoucke2011improving,jaderberg2014speeding,sandler2018mobilenetv2}. By matching the output of a large neural network, or \emph{teacher}, with that of a small lightweight model, or \emph{student}, KD aims at transferring the performance of state-of-the-art networks to small architectures~\cite{beyer2021knowledge}. Surprisingly, it has recently been discovered that, even when both models have the same capacity, sometimes the student can perform better than its teacher~\cite{furlanello2018born,bagherinezhad2018label,dong2019distillation}. This effect has been attributed to the existence of additional hidden information about the learned representations of the teacher in its outputs (e.g. similarity between classes) which the student can exploit more than the original labels. This information is usually referred to as \emph{dark knowledge}~\cite{hinton2015distilling,furlanello2018born,yang2018knowledge}.

In the context of adversarial robustness~\cite{optimism_proc_ieee}, though, KD has been mostly proposed just as a way to create lightweight robust models that can be safely deployed in mobile devices~\cite{goldblum2020adversarially,zi2021revisiting,shao2021and}. In particular, it has been observed that distilling knowledge from an adversarially trained large teacher can successfully transfer its robustness to a smaller student. Nevertheless, to the best of our knowledge, KD between large models of the same capacity has not been studied thoroughly before, so it still remains an open question to know if distillation can be used to improve robustness performance rather than just compressing. 



\begin{table}[t]
    \caption{Knowledge distillation using the proposed AKD framework can boost the clean and the robust performance of the student model. The effect is more noticeable on bigger models. Results for $L_\infty$ perturbations of size $\varepsilon=8/255$. Std: standard training. AT: adversarial training~\cite{madry2017towards}. AT+: adversarial training using improvements from~\cite{rebuffi2021fixing}}
    \label{tab:main_results}
    \centering
    \begin{tabular}{c|cc|cc}
         \toprule
          Dataset & Student & Teacher & Clean & AutoAttack~\cite{croce2020reliable} \\
         \hline
         \multirow{5}{5em}{CIFAR10} & ResNet-18 (AT) & No teacher & 84.22 & 46.99 \\
         & ResNet-18 & ResNet-18 (Std.) & $\textbf{86.28}$ & $44.63$ \\
         & ResNet-18 & ResNet-18 (AT) & $82.70$ & $\textbf{47.66}$ \\
         \cline{2-5}
         & WideResNet-28 (AT+) & No teacher & 87.78 & 58.18 \\
         & WideResNet-28 & WideResNet-28 (AT+) & \textbf{88.40} & \textbf{61.23}\\
         \midrule
         \multirow{5}{5em}{CIFAR100} & ResNet-18 (AT) & No teacher & 56.91 & 19.67 \\
         & ResNet-18 & ResNet-18 (Std.) & $\textbf{62.09}$ & $19.56$ \\
         & ResNet-18 & ResNet-18 (AT) & $56.87$ & $\textbf{24.61}$ \\
         \cline{2-5}
         & WideResNet-28 (AT+) & No teacher & 61.98 & 29.58 \\
         & WideResNet-28 & WideResNet-28 (AT+) & \textbf{62.18} & \textbf{30.67}\\
         \bottomrule
    \end{tabular}
\end{table}

For this reason, in this work, we present a thorough analysis 
to gauge the ability of KD as a way to improve robustness. Specifically, we propose a new form of adversarial knowledge distillation (AKD) which can be effectively used to improve the performance of a robust model, both in terms of robustness and clean accuracy depending on the robustness of the teacher (see Tab.~\ref{tab:main_results}). Furthermore, we provide a rigorous ablation study on AKD, and other robust KD methods, and shed new light on their key components. Our main goal is to understand how different modes of KD can influence the student's performance, with the aim to provide actionable insights on how to use it.

Overall, the main discoveries of our work are:
\begin{itemize}
    \item[$\bullet$] KD can be used to boost the performance of adversarially robust models as a simple plugin method to improve robustness and/or clean accuracy of state-of-the-art robust models. Interestingly, linearly mixing clean labels with the adversarial outputs of a pretrained teacher is sufficient to do so.
    \item[$\bullet$] Distilling from an adversarially trained teacher tends to increase robustness, while distilling from a standardly train model increases clean accuracy. Surprisingly, models trained adversarially using small $\varepsilon$ attacks~\cite{optimism_proc_ieee} can further improve clean accuracy while also transferring some robustness.
    \item[$\bullet$] Early-stopped teachers maximize robustness on the student model. The stopping point for pretraining that maximizes post-distillation performance, is a network- and dataset-specific hyperparameter.
    \item[$\bullet$] Distilling knowledge from an ensemble of models can further improve performance, even when the individual models are only weakly robust. Combined ensembles of standard and robust teacher models can be used to achieve a wider range of clean and robust performance trade-offs.
    \item[$\bullet$] Finally, we show that AKD has different effects on different training samples~\cite{liu2021impact,dong2021data}. It reduces the confidence of easy-to-learn instances, and increases it on the harder ones. This allows the student to learn from samples that adversarial training alone has a hard time to even memorize~\cite{arpit2017closer}.
\end{itemize}

We provide deep analysis and solid empirical evidence for all of these discoveries, and show their wide applicability. This further motivates the use of KD, not only to compress big networks, but to also increase their robustness performance. We believe that our insights can have a large impact on future work on KD for adversarial robustness, as they give easy to follow guidelines to apply these methods on different contexts.  

\section{Adversarial Knowledge Distillation}

KD consists on training a student model $f_S:\R^d\to\R^c$ to mimic the outputs of a teacher model $f_T:\R^d\to\R^c$. To that end, most KD methods~\cite{hinton2015distilling,romero2014fitnets,zagoruyko2016paying,chebotar2016distilling} append a matching function to the loss which encourages the final representations of $f_S$ and $f_T$ to be close at every sample $\bm x\in\R^d$ with associated label $y\in\{0,1\}^c$. In other words, Clean Knowledge Distillation (CKD) methods optimize the loss
\begin{equation}
    \text{CKD}(\bm x, y) = (1-\lambda)\text{CE}(f_S(\bm x), y) + \lambda \text{KL}(f_S(\bm x), f_T(\bm x))
\end{equation}
where $f_S(\bm x)$ and $f_T(\bm x)$ are the probability outputs of the student and the teacher models, CE denotes the cross-entropy loss, KL represents the Kullback-Leibler divergence, and $\lambda\in[0,1]$ controls the strength of the regularization.


While this loss is effective in transferring clean performance, it does not seem to be able to transfer robustness against adversarial examples~\cite{goldblum2020adversarially}
\begin{equation*}
    \bm x' = \arg \max_{\|\bm x' - \bm x\| \leq \varepsilon} \text{CE}(f(\bm x'), y),
\end{equation*}
which can be generally obtained using Projected Gradient Descent (PGD)~\cite{madry2017towards}. 

For this reason, recent works have proposed different \emph{robust knowledge distillation} (RKD) alternatives to address this problem~\cite{goldblum2020adversarially,zi2021revisiting,zhu2021reliable,shao2021and}. Specifically, Adversarially Robust Distillation (ARD)~\cite{goldblum2020adversarially} proposes to add a regularization term that encourages robustness by matching $f_S$ and $f_T$ at the output of adversarial and clean examples, i.e., 
\begin{equation}
    \text{ARD}(\bm x, y) = (1 - \lambda)\text{CE}(f_S(\bm x), y) + \lambda\text{KL}(f_S(\bm x'), f_T(\bm x)).
    \label{eq:ard}
\end{equation}
Meanwhile, Robust Soft Label Adversarial Distillation (RSLAD)~\cite{zi2021revisiting} avoids the use of clean labels and focuses only on matching the models' outputs, i.e.,
\begin{equation}
    \text{RSLAD}(\bm x) = (1 - \lambda)\text{KL}(f_S(\bm x), f_T(\bm x)) + \lambda\text{KL}(f_S(\bm x'), f_T(\bm x)).
    \label{eq:rslad}
\end{equation}

RSLAD and ARD are effective at distilling robust knowledge from large models to smaller ones, but they are not designed to improve the robust performance of a standalone model. With this in mind, and inspired by the success of self-distillation~\cite{dong2019distillation}, we propose Adversarial Knowledge Distillation (AKD), a new distillation method that mixes the outputs of a teacher network on adversarial examples with the clean labels while performing adversarial training. In particular, AKD solves the following objective
\begin{equation}
\label{eq:akd}
    \text{AKD}(\bm x, y) = \text{CE}(f_S(\bm x'), \alpha f_T(\bm x') + (1-\alpha)y)
\end{equation}
where $\alpha\in[0,1]$ controls the mixing of the distilled labels and the original ones.

AKD circumvents two issues of the prior methods: i) By distilling knowledge of the teacher at the adversarial regions, it ensures that the distillation process drives the student model to exactly match the function outputs of the teacher~\cite{beyer2021knowledge}. ii) By mixing the adversarial outputs and the clean labels, it allows the student model to compensate the variable quality of the distilled labels during training. As we will see, this is an important feature in the context of adversarial training where the existence of hard-to-learn samples and label noise can severely hamper the final performance of a model~\cite{liu2021impact,dong2021data}. Furthermore, note that label mixing changes the optimum of the optimization to $\alpha f_T(\bm x') + (1-\alpha)y$, heavily penalizing model over-confidence for $\alpha > 0$. This can help reduce the model Lipschitz constant around the data points (i.e. maximum gradient with respect to the input), thus making the student more robust~\cite{cisse2017parseval}.

Through the rest of this paper we will fundamentally study AKD to show that KD is an effective tool to effectively improve robustness in conjunction with state-of-the-art methods: In this sense, we will mostly analyze how different design choices in AKD affect robustness, but by the end of this work, we will compare AKD to other robust distillation methods and show that all our insights also generalize to other RKD techniques, albeit with a slightly lower performance.

\section{Teacher training trade-offs}
\label{sec:teacher_tradeoffs}
In this section, we motivate and analyze different parameter and optimization choices on the teacher model and explore their impact in the context of RKD. In what follows, and to simplify the complexity of our experimental settings\footnote{The code will be open-sourced upon acceptance.}, we will only consider the case where the student and teacher model share the same architecture. Specifically, we will mostly use a ResNet-18~\cite{he2016deep} in the majority of our experiments. The training details of all our experiments can be found in the supplementary material. However, as a general rule, we train our teachers using stochastic gradient descent (SGD) with an exponential learning rate decay for 50 epochs, unless early-stopped. Afterwards, we train the students using the AKD loss function in \eqref{eq:akd} using SGD with a cyclic learning rate decay for 50 epochs. In both cases, the initial learning rate is 0.1 and the batch size is 256. Unless stated otherwise, we use PGD-7 and AutoAttack L$_{\infty}$ perturbations of size $\varepsilon=8/255$ to perform adversarial training and evaluate robustness, respectively.

At this point, we would like to highlight that the adversarial robustness problem is inherently a multi-task objective: That is, in light of the robustness-accuracy trade-off~\cite{tsipras2018robustness}, it seems difficult to maximize robustness without a certain drop in clean accuracy. Our goal with the following experiments is thus to emphasize the setups under which we can use AKD, and RKD in general, to push the pareto frontier of adversarial robustness.


\subsection{Early stopping}
\label{sec:es}

\begin{figure}[t]
    \centering
    \includegraphics[width=0.45\textwidth]{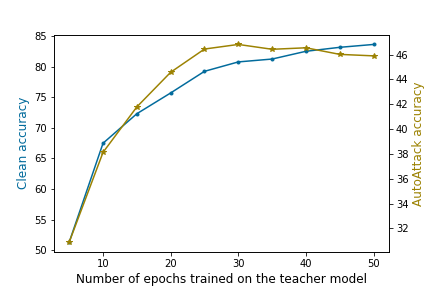}
    \caption{Effect of early stopping the teacher on the AKD method. The train the teacher model using adversarial training. We use the ResNet-18 architecture for the teacher and the student, on the CIFAR10 dataset~\cite{krizhevsky2009learning}.}
    \label{fig:akd_es}
\end{figure}

It has been shown that using early-stopped teachers in CKD can lead to improved accuracies on the student model. This improvement has been attributed to the mismatched capacity between the teacher and the student model~\cite{cho2019efficacy,yang2018knowledge}. Nevertheless, we now show, that using early-stopped teachers also presents advantages in RKD between models of same capacity. 
To illustrate the benefits of early stopping, we compare in Figure \ref{fig:akd_es} how the performance of the student model varies depending on when we early-stopped its teacher. Here, the student is trained with AKD using $\alpha=1.0$ to avoid any confounding effect from label mixing. Meanwhile, the teacher is simply adversarially trained with our standard settings. From this plot, we can observe there is an optimal early stopping point where the student robustness is maximized. In contrast, we also see that distilling from an early-stopped teacher reduces the student clean accuracy.

We believe this phenomenon is due to multiple compounded effects. First, note that in the later stages of adversarial training the teacher network is memorizing rather than learning new features~\cite{arpit2017closer,rahaman2019spectral}. This may hinder the transfer of robustness to the student network~\cite{liu2021impact,dong2021data}. Second, early stopping the teacher allows to distill from a model which is less confident of its outputs thus maximizing the available dark knowledge~\cite{mirzadeh2019improved}. And finally, using an early-stopped adversarially trained teacher can prevent negative effects stemming from catastrophic or robust overfitting~\cite{andriushchenko2020understanding,rice2020overfitting}.



\subsection{Label mixing}
\label{sec:alpha}

It has been recently observed that adversarial training consistently introduces a certain degree of label noise that can hurt its performance~\cite{sanyal2020benign}, while standard training might tend to memorize certain parts of the training set~\cite{arpit2017closer,rahaman2019spectral}. Therefore, we now study the effect of using label mixing, controlled by the hyperparameter $\alpha$, as a way to correct possible teacher mistakes and thus improve the final performance of AKD. 

\begin{figure}[t]
    \centering
    \includegraphics[width=0.45\textwidth]{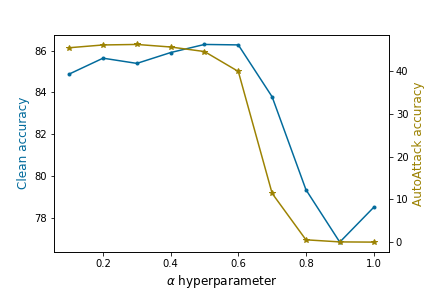}
    \includegraphics[width=0.45\textwidth]{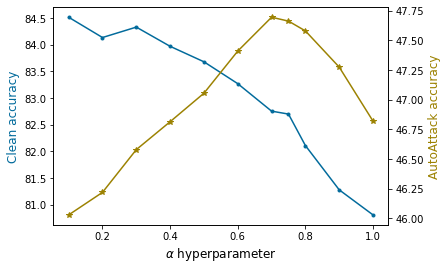}
    \caption{Effect of label mixing on the AKD method. We used a standardly (left) and an adversarially (right) trained ResNet-18 for the teacher model, and a ResNet-18 for the student, on the CIFAR10 dataset.}
    \label{fig:akd_alpha}
\end{figure}

We compare in Figure \ref{fig:akd_alpha} the performance of the student model when we change the value of $\alpha$. Note that increasing/decreasing $\alpha$ is a way to control our confidence on the teacher outputs, and thus control the impact of KD.  We consider a standardly trained teacher and an adversarially trained teacher, early-stopped at epochs 20 and 30, respectively. In these plots, we observe that label mixing is especially helpful when we use a standardly trained teacher, as it avoids transferring the teacher susceptibility to adversarial perturbations while transferring some of its clean performance. For adversarially trained teachers, we observe that the student robustness is maximized with a high $\alpha$ value. Thus, when the teacher is robust, its outputs transfer robustness much more effectively, but still, it is better to slightly mix the original labels to correct the samples that the robust teacher struggles to learn~\cite{liu2021impact,dong2021data}. Similar results can be obtained for teachers of a higher capacity, and for other methods like RSLAD (see supplementary material).

\subsection{Teacher robustness}
\label{sec:teacher_robustness}

We now study the impact of the robustness of the teacher on the student performance. Intuitively, this should be the most important factor, since we partly distill the teacher outputs to train the student model with AKD. Thus, if the teacher is not robust enough, matching its outputs by distillation will result in the student model being weakly robust as well.

In Tab.~\ref{tab:akd_teacher_training}, we present a comparison of the student performance when we distill from differently trained teachers using AKD. We compare between teachers trained using standard and adversarial training, and for the latter we distinguish between networks trained using either PGD with 7 iterations (PGD-7)~\cite{madry2017towards} or Fast-FGSM (FFGSM)~\cite{wong2020fast} adversarial training techniques which allow to obtain slightly less robust models in a fraction of the time. In these experiments, we optimize the early stopping epoch of the teacher, and the label mixing parameter, to maximize robust performance when the teacher is trained adversarially, and clean performance otherwise\footnote{The results for more conservative trade-offs can be found in the supp. material}.
We consistently observe that standardly trained teachers tend to boost the clean performance of the student while preserving robustness, while adversarially trained teachers tend to increase adversarial performance, at the cost of some clean accuracy. Moreover, we see that despite not boosting the robust performance as much as the PGD-7 models, the weaker FFGSM models tend to achieve better trade-offs between robustness and accuracy. 

Surprisingly, we find that all instances of AKD achieve even better results on the more challenging CIFAR100 dataset~\cite{krizhevsky2009learning}. We believe this is a result of the hierarchical class structure of the CIFAR100 dataset, which results in many classes being very similar to others. This information can be transferred efficiently to the student using KD, since the distilled probabilities of similar classes will tend to be more correlated~\cite{hinton2015distilling}.

\begin{table}[t]
    \centering
    \caption{Impact of differently trained teachers on the student performance on the AKD method. For all adversarial attacks, we use $L_\infty$ norm of size $\varepsilon=8/255$.}
    \label{tab:akd_teacher_training}
    \begin{tabular}{ccc|cc}
        \toprule
          Dataset & Student & Teacher & Clean & AutoAttack \\
         \hline
          \multirow{4}{5em}{CIFAR10} & ResNet-18 (AT) & No teacher & $84.22$ & $46.99$ \\
          & ResNet-18 & ResNet-18 (Std.) & $\textbf{86.28}$ & $44.63$ \\
          & ResNet-18 & ResNet-18 (AT$_{\text{FFGSM}}$) & $83.80$ & $47.05$ \\
         & ResNet-18 & ResNet-18 (AT$_{\text{PGD-7}}$) & $82.70$ & $\textbf{47.66}$ \\
         \midrule
         \multirow{4}{5em}{CIFAR100} & ResNet-18 (AT) & No teacher & $56.91$ & $19.67$ \\
          & ResNet-18 & ResNet-18 (Std.) & $\textbf{62.09}$ & $19.56$ \\
         & ResNet-18 & ResNet-18 (AT$_{\text{FFGSM}}$) & $58.38$ & $23.44$ \\
         & ResNet-18 & ResNet-18 (AT$_{\text{PGD-7}}$) & $56.87$ & $\textbf{24.61}$  \\
         \bottomrule
    \end{tabular}
\end{table}

Finally, inspired by works from the adversarial robustness literature that claim that adversarial training with very small perturbations can improve generalization~\cite{xie2020adversarial,bochkovskiy2020yolov4}, we also look at the effect of the adversarial perturbation region size in RKD.
We compare how the performance of the student varies depending on the $\varepsilon$ size of the attack used to train the teacher model adversarially (see Tab.~\ref{tab:small_eps}). We show that, for a small $\varepsilon$, we can get better accuracy and robustness than when we used standard training. This increase in generalization can be transferred to the student, even when the student is still trained using perturbations with the standard $\varepsilon$ value of 8/255 used in CIFAR datasets. Similarly, we show that using a large $\varepsilon$ gives worse results, probably because the perturbations are too large and can be perceptible~\cite{sharma2017attacking}.

\begin{table}[t]
    \centering
    \caption{Effect of using different $L_\infty$ ball constraints to adversarially train the teacher model on the AKD method. We use the ResNet-18 architecture for both teacher and student models.}
    \label{tab:small_eps}
    \begin{tabular}{cc|cc}
        \toprule
          Dataset & $L_\infty$ ball size ($\varepsilon$) & Clean & AutoAttack \\
         \hline
          \multirow{4}{5em}{CIFAR10} & 0 & $86.28$ & $44.63$ \\
          & 2/255 & $\textbf{86.59}$ & $45.42$ \\
          & 8/255 & $82.70$ & $\textbf{47.66}$ \\
          & 16/255 & $82.15$ & $47.25$ \\
          \midrule
           \multirow{3}{5em}{CIFAR100} & 0 & $61.90$ & $21.23$ \\
           & 2/255 & $\textbf{62.09}$ & $22.12$ \\
           & 8/255 & $56.87$ & $\textbf{24.61}$ \\
           \bottomrule
    \end{tabular}
\end{table}

\subsection{Ensemble of teachers}
\label{sec:ensemble}

Until this point, we have only trained a single teacher. However, we now show that using an ensemble of similarly trained teachers, which only differ on their random initialization, can distill robustness even further. This technique is commonly used in CKD~\cite{chebotar2016distilling,cui2017knowledge,freitag2017ensemble} to improve clean accuracy, but it is unclear if it should work to increase robustness as well. Especially, since using an ensemble of models does not seem to improve robustness on its own (see supplementary material).

Tab.~\ref{tab:akd_ensemble} shows an equivalent analysis to Subsection \ref{sec:teacher_robustness}, but using an ensemble of similarly trained teachers rather than a single model. To that end, we generalize the AKD objective for an ensemble of teachers $\{f_{T_i}\}_{i=1}^M$ as
\begin{equation}
    \text{AKD}(\bm x, y) = \text{CE}\left(f_S(\bm x'), \; \alpha \sum_{i=1}^M \beta_i f_{T_i}(\bm x')+ (1-\alpha)y \right)
\end{equation}
where $M$ is the number of networks in the ensemble and $\beta_i$ controls the mixing of all the individual teachers ($\beta_i \geq 0$, $\sum_i \beta_i = 1$). Since all the teachers are similarly trained, we use $\beta_i = 1/M$.

Surprisingly, we find that using an ensemble of teachers does not only improve clean accuracy but also the robust performance for most training configurations. In particular, it greatly improves the accuracy when the teachers are standardly trained, and it is effective in further improving robustness when the teachers are adversarially trained. We highlight that, assuming an increased computational cost due to training all the teacher models from scratch, it is much more preferable to use an ensemble of models adversarially trained with FFGSM rather than a single model trained with PGD-7 since both options have similar cost\footnote{For the computational cost comparison, we use that AT$_{\text{FFGSM}}$ costs 2 times more than standard training, and AT$_{\text{PGD-7}}$ costs 8 times more}. At the same time, compared to the cost of computing PGD-7 adversarial perturbations to train with AKD, using an ensemble of standardly trained models can be very cost-effective.

\begin{table}[t]
    \centering
    \caption{Effect of using an ensemble for the teacher model on the AKD method. We use the ResNet-18 architecture for both student and teacher models. For all adversarial attacks, we use $L_{\infty}$ norm of size 8/255.}
    \label{tab:akd_ensemble}
    \begin{tabular}{cc|cc|cc}
        \toprule
         \multicolumn{2}{c|}{} & \multicolumn{2}{c|}{1 model} & \multicolumn{2}{c}{4 models}\\
    	    \cline{3-6}
          Dataset & Teacher training & Clean & AutoAttack & Clean & AutoAttack \\
         \hline
          \multirow{3}{5em}{CIFAR10} & Std. & $\textbf{86.28}$ & $44.63$ & $\textbf{87.27}$ & $43.36$ \\
         & AT$_{\text{FFGSM}}$ & $83.80$ & $47.05$ & $84.05$ & $47.51$ \\
         & AT$_{\text{PGD-7}}$ & $82.70$ & $\textbf{47.66}$ & $82.96$ & $\textbf{47.96}$ \\
         \midrule
          \multirow{3}{5em}{CIFAR100} & Std. & $\textbf{62.09}$ & $19.56$ & $\textbf{63.72}$ & $20.85$ \\
         & AT$_{\text{FFGSM}}$ & $58.38$ & $23.44$ & $58.79$ & $24.64$ \\
         & AT$_{\text{PGD-7}}$ & $56.87$ & $\textbf{24.61}$ & $58.34$ & $\textbf{24.93}$ \\
         \bottomrule
    \end{tabular}
\end{table}

\begin{figure}[t]
    \centering
    \includegraphics[width=0.48\linewidth]{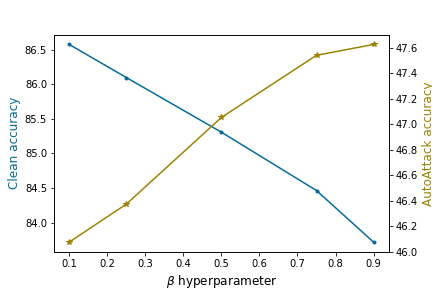}
    \caption{Effect of varying the mixing of two standardly trained and adversarially trained teachers on the AKD method. We use the ResNet-18 architecture for both the teacher and student models, used with the CIFAR10 dataset.}
    \label{fig:beta_effect}
\end{figure}

Finally, one of the main advantages of using an ensemble of teacher models is that we are not constrained to use one type of training for all models. This allows us to have much more control when choosing the trade-off, since we can use different $\beta_i$ values for each type of model. We show in Figure~\ref{fig:beta_effect} the performance of the student when mixing two standardly trained and adversarially trained teacher models, where $\beta$ determines the mixing weight of the adversarially trained teacher model. We observe that the change in performance as we increase $\beta$ is quite smooth, allowing us to obtain student models with possibly more desirable clean and robust performance trade-offs than when all the teachers are standardly or adversarially trained (Tab.~\ref{tab:akd_ensemble}).

\subsection{Insights about other robust distillation methods}
\label{sec:comparison_rslad}

So far, we have found and proposed several tricks and guidelines to improve the performance of a student model using AKD.
However, we will now show that these insights are not exclusive to AKD, and translate also to other RKD methods.

Particularly, we apply the previous improvements to RSLAD~\cite{zi2021revisiting}\footnote{We focus on this method as it seems to outperform ARD~\cite{goldblum2020adversarially}}. Note that in the original formulation RSLAD does not use the clean labels. Hence, in these experiments we use a slight generalization of RSLAD in which we also mix the original labels alongside the teacher outputs. 

Strikingly, we find in Table~\ref{tab:other_distill} that RSLAD does also benefit from early stopping, mixing the distilled labels, and using an ensemble instead of a single model. Furthermore, we see that RSLAD can also enhance clean and robust performance between models of the same capacity depending on the robustness of the teacher. When compared with RSLAD, we find our AKD method has comparable performance. However we tend to find AKD easier to tune thanks to its simplicity, thus allowing us to cover a larger part of the pareto frontier. As a general rule of thumb, though, we find that all insights from AKD seem to translate to other RKD methods, thus making KD a competitive technique to boost the robust performance of any model.

\begin{table}[t]
    \centering
    \caption{Effect of early stopping, label mixing and using different $L_\infty$ ball constraints to adversarially train the teacher for different RKD methods. ES $\&$ LM: teacher early stopping and label mixing}
    \label{tab:other_distill}
    \begin{tabular}{cccc|cc}
        \toprule
          KD method & $\varepsilon$ & ES \& LM & Ensemble & Clean & AutoAttack \\
         \hline
         None & 8/255 & No & No & $84.22$ & $46.99$ \\
         \hline
          & 0 & No & No & 85.47 & 46.11 \\
          & 0 & Yes & No & $85.47$ & $46.39$ \\
         RSLAD & 0 & Yes & Yes & $85.82$ & $46.49$ \\
          & 2/255 & Yes & No & $85.40$ & $46.51$ \\
          & 8/255 & Yes & No & $82.60$ & $47.31$ \\
         \hline
          & 0 & Yes & Yes & $\textbf{86.91}$ & $45.64$ \\
          AKD & 2/255 & Yes & No & $86.25$ & $46.13$ \\
          & 8/255 & Yes & No & $82.70$ & $\textbf{47.66}$ \\
          \bottomrule
         
    \end{tabular}
\end{table}

All in all, we have presented different ways to train a teacher model and distill its knowledge using RKD and found that using early-stopping, label mixing and an ensemble of teachers results in better overall performance. Particularly, we found that student accuracy and robustness can be maximized by using small-$\varepsilon$ and standard adversarial training, respectively. And finally, we have shown that training the teacher with weak attacks is a cost-effective solution in terms of RKD performance. In the next section, we will thus apply all these insights with the aim to push even further the robustness of state-of-the-art models.

\section{Performance Analysis of AKD}

\subsection{Robustness}
\label{sec:sota_comparison}

We have shown that AKD can be successfully used as a simple plugin method to boost the performance of a naive adversarially trained model. With the motivation of generalizing these results to more sophisticated robust models, we combine AKD with the current state-of-the-art adversarial training techniques~\cite{rebuffi2021fixing} and show that it can be effectively used to improve over the state-of-the-art performance.

To maximize the robustness of the model, the authors in~\cite{rebuffi2021fixing} use model weight averaging~\cite{izmailov2018averaging}, extra data sampled from generative models~\cite{ho2020denoising} and data augmentation techniques like Cutout~\cite{devries2017improved}. We use these optimization blocks with the default hyperparameters given in~\cite{rebuffi2021fixing} for training the teacher, using adversarial training, and the student, using our AKD method. Both teacher and student use the WideResNet-28~\cite{he2016deep,zagoruyko2016wide} with swish activations~\cite{hendrycks2016gaussian} architecture, as proposed in~\cite{rebuffi2021fixing}. We show in Table~\ref{tab:sota} that AKD, even without using label mixing, significantly boosts the student performance compared to its teacher, the current state-of-the-art model in robustness.
We also measure the effect of some of our previously discussed design choices: label mixing and teacher early-stopping. On the one hand, we find that early stopping is much less effective. We believe this is due to the data augmentation proposed by the authors, which significantly delays when the network starts memorizing training samples. On the other hand, we show that, even when the teacher performs quite well, label mixing can further improve the clean and robust performance of the model compared to just using the distilled labels, which is consistent with our previous results. 

\begin{table}[t]
    \centering
    \caption{Comparison between the teacher trained using the current state-of-the-art method (AT+)~\cite{rebuffi2021fixing} and the student trained using AKD. The student and the teacher use the WideResNet-28~\cite{he2016deep,zagoruyko2016wide} with swish activations~\cite{hendrycks2016gaussian} architecture.}
    \label{tab:sota}
    \begin{tabular}{cccc|ccc}
        \toprule
          Dataset & Training type & $\alpha$ & ES & Clean & PGD-20 & AutoAttack \\
         \hline
         \multirow{4}{5em}{CIFAR10} & AT+ & --- & No & 87.78 & 61.60 & 58.18 \\
         & AKD & 1.0 & No & 88.09 & 62.58 & 60.55 \\
         & AKD & 1.0 & ep. 100 & 87.41 & 61.47 & 59.97\\
         & AKD & 0.8 & No & \textbf{88.40} & \textbf{63.19} & \textbf{61.23} \\
         \midrule
         \multirow{3}{5em}{CIFAR100} & AT+ & --- & No & 61.98 & 31.88 & 29.58 \\
         & AKD & 1.0 & No & 61.86 & \textbf{33.43} & 30.58 \\
         & AKD & 0.8 & No & \textbf{62.18} & 33.19 & \textbf{30.67} \\
         \bottomrule
    \end{tabular}
\end{table}

\subsection{Training trajectories}
\label{sec:trajectories}

We finally provide an empirical study of the effect that AKD has on the functional properties of the student model. In particular, we find that using AKD modifies the calibration of the student model, mostly by modifying the training trajectory on hard-to-learn samples.

To test this behaviour, we first compare the evolution of the average entropy of the output probabilities of different models $H=\mathbb{E}_{\bm x}[-\sum_i f_S(\bm x)_i\log f_S(\bm x)_i]$, which measures how uncertain the model is on its predictions during training. We compute the probabilities on all the training data, and train using a ResNet-18 on CIFAR10. As we can see, AKD forces the student to learn solutions while being less over-confident than its teacher. This observation agrees with recent reports that suggest that CKD students are also better calibrated and less over-confident than standard models~\cite{muller2019does}.


\begin{figure}[t]
    \centering
    \includegraphics[width=0.5\textwidth]{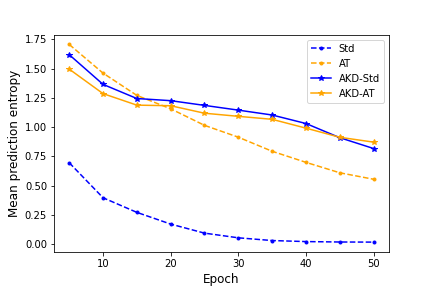}
    \caption{Mean entropy value of different optimization methods as the training progresses. The architecture is ResNet-18, used with the CIFAR10 dataset. AKD-Std and AKD-AT use AKD with a standardly trained and adversarially trained teacher, respectively.}
    \label{fig:entropy_kd}
\end{figure}

It is widely known that deep neural networks do not use all training samples in the same way, and that certain training samples are harder to fit~\cite{arpit2017closer}. What is more, in the context of adversarial robustness, it has been recently observed that the impact of harder-to-learn instances is even more noticeable than for standard models~\cite{sanyal2020benign,liu2021impact,dong2021data}. In this sense, it is expected that KD will have a stronger effect on those examples that can better  exploit the \emph{dark knowledge} of the teacher.

In order to test this hypothesis, we train our models on a binary-class dataset composed of all CIFAR10 samples that belong to the two first classes: \emph{car} and \emph{airplane}. Moreover, and following the practice in~\cite{liu2021impact,dong2021data}, we rank the difficulty of learning each training and test sample based on the score
\begin{equation}
\label{eq:difficulty}
    \mathcal{S}(\bm x)=\dfrac{1}{K} \sum_{k = 1}^K \text{CE}(f_k(\bm x), y),
\end{equation}
where $f_k$ denotes the state of the neural network after $k$ epochs of training, and $K$ denotes the total number of training epochs.


Ranking the samples allows us to determine when does AKD improve performance. For each sample, we measure the improvement by recording the probability the model gives to the correct class\footnote{Well-calibrated models should assign high probabilities to the correct class, but not be too overconfident} at the end of every epoch. We collect these values into a vector of probabilities $\bm p_S(\bm x)\in[0,1]^K$ in the case of the student, and $\bm p_T(\bm x)\in[0,1]^K$ in the case of the teacher. These vectors summarize the functional training trajectory of the models on a given sample $\bm x$, which means that their inner product measures the differences in their training dynamics.

\begin{figure}[t]
    \centering
    \includegraphics[width=0.45\textwidth]{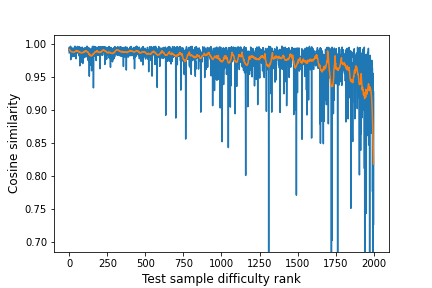}
    \includegraphics[width=0.45\textwidth]{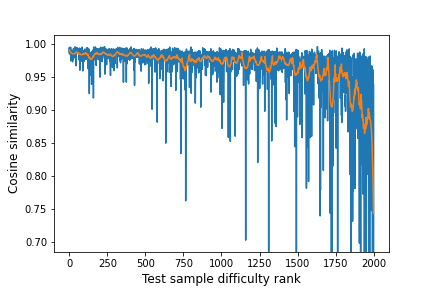}
    \caption{Cosine similarity when using AKD with an adversarially trained teacher when evaluated on natural (left) and adversarial (right) images. The blue line shows the true values, and the orange line a smoothed-out version.}
    \label{fig:all_at_akd_cossim}
\end{figure}

\begin{figure}[t]
    \centering
    \subfloat[$p_{T,K}$]{
        \centering
        \includegraphics[width=0.45\textwidth]{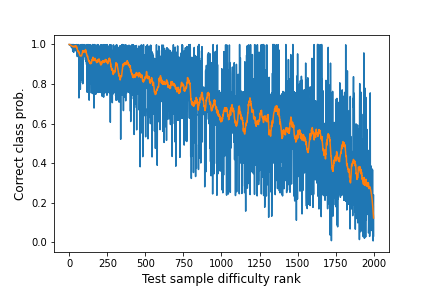}
    }
    \subfloat[$p_{S,K} - p_{T,K}$]{
        \centering
        \includegraphics[width=0.45\textwidth]{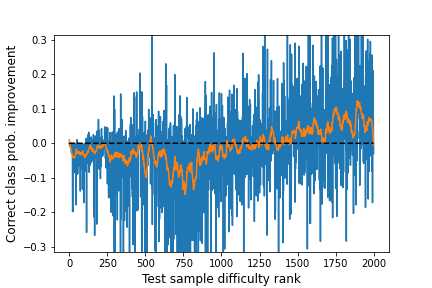}
    }
    \caption{Correct class probability of the adversarially trained teacher (left) and the relative improvement of the student (right) tested on adversarial images. The blue line shows the true values, and the orange line a smoothed-out version.}
    \label{fig:all_at_akd_distil}
\end{figure}

Figure \ref{fig:all_at_akd_cossim} shows precisely these differences on all natural and adversarial test samples, ordered in terms of learning difficulty\footnote{We provide results for a standardly trained teacher in the supplementary material}. As we can see, those samples which are easier to learn mostly follow the same trajectory on the student and teacher models, while the training trajectories clearly differ on hard examples, especially if they are adversarial. Interestingly, these differences in trajectory tend to have a strong calibration effect. As shown in Figure \ref{fig:all_at_akd_distil}, the student model tends to assign higher probabilities on hard samples than the teacher, while it also tends to slightly reduce the confidence on the very easy examples. This effect is especially beneficial in terms of robustness, as it is consistent with the fact that difficult samples and label noise are more harmful to robustness than clean performance~\cite{sanyal2020benign,liu2021impact,dong2021data}. We believe this ability to improve generalization to difficult-to-learn regions can make KD very promising for multiple applications that require better performance in out-of-distribution data.

\section{Related work}

As we have discussed in our work, we found that may techniques and ideas from KD are not well-explored by RKD~\cite{goldblum2020adversarially,zi2021revisiting,shao2021and} methods, which mostly focus on compressing robust models. In contrast, our proposed AKD method and some of our design choices are inspired by CKD methods that are focused on improving the model performance. Label mixing is inspired from self-distillation~\cite{dong2019distillation}, in which the outputs of the teacher are mixed with the clean labels. Moreover, there are other CKD works that proposed to use an ensemble of teachers to improve clean performance~\cite{chebotar2016distilling,cui2017knowledge,freitag2017ensemble}, which we show can also increase robustness when used on RKD methods. Our method also contrasts with other CKD~\cite{hinton2015distilling} and RKD~\cite{goldblum2020adversarially,zi2021revisiting,shao2021and} methods in that function matching happens implicitly in the optimization instead of as an additional regularization term. Particularly, we differ from RKD methods in that use function matching exclusively in the adversarial region, inspired by~\cite{beyer2021knowledge,dong2019distillation}. The increased improvement of our method on the more complex CIFAR100 dataset is consistent with the claim that KD encodes the semantic similarity between classes~\cite{hinton2015distilling,furlanello2018born,yang2018knowledge}. Moreover, the fact that we obtain better performance using early-stopping has been theorized in the context of memorization, where it is claimed that neural networks learn some of the most important features very early and rely on memorization in the later steps of the training~\cite{arpit2017closer,dong2019distillation}. 

In the context of adversarial robustness, we find that AKD penalizes overconfidence on easier examples and improves performance on difficult-to-learn regions. This is consistent with recent adversarial literature, that shows that robustness is difficult to achieve due to the network not being able to memorize harder instances or being affected by label noise~\cite{sanyal2020benign,liu2021impact,dong2021data,rahaman2019spectral}, and that networks with reduced Lipschitz constant are more robust~\cite{cisse2017parseval}. In this work, we show that AKD can complement the current state-of-the-art methodology, that combines training with adversarial examples~\cite{madry2017towards,zhang2019theoretically}, model weight averaging~\cite{izmailov2018averaging}, extra data sampled from generative models~\cite{ho2020denoising} and data augmentation techniques like Cutout~\cite{devries2017improved}.

\section{Conclusion}

In this work, we presented a thorough analysis that shows that KD can be used to boost the model robustness, surpassing state-of-the-art performance. We proposed a new robust KD algorithm, AKD, and gave guidelines to use early stopping, label mixing, ensembling, and small-epsilon adversarial training to boost the student model performance. Finally, we analyzed how KD affects the training trajectories of different samples, and found that it calibrates the model and improves its performance in difficult-to-learn regions. 

In the future, we think our robustness results could be  improved improved even further by using an ensemble of teachers, small-$\varepsilon$ adversarial training and further tuning the label mixing parameter. Moreover, it would be interesting to have results on larger architectures and more datasets.

%
%
\bibliographystyle{splncs04}
\bibliography{references}

\clearpage
\appendix
\setcounter{page}{1}
\setcounter{table}{0}
\setcounter{figure}{0}
\renewcommand{\thefigure}{S\arabic{figure}}
\renewcommand{\thetable}{S\arabic{table}}

\include{appendix}

\bibliographystylesupp{splncs04}
\bibliographysupp{references}

\end{document}

%% file: appendix.tex
\section{Details on the experimental setup}


\subsection{Architectures and training}

For all our experiments in Section \ref{sec:teacher_tradeoffs} and Section \ref{sec:trajectories}, we use the ResNet-18 architecture~\cite{he2016deep} for both teacher and student models. For Section \ref{sec:sota_comparison}, we use the WideResNet-28~\cite{he2016deep,zagoruyko2016wide} with swish activations~\cite{hendrycks2016gaussian} architecture, as proposed in~\cite{rebuffi2021fixing}, for both teacher and student models. 

We use the same training parameters for both CIFAR10 and CIFAR100 datasets~\cite{krizhevsky2009learning}.
For both student and teacher models, we use stochastic gradient (SGD) for 50 epochs, with 0.1 as the initial learning rate.
Because we found that early-stopping the teacher is beneficial when using AKD, we train it using exponential learning rate decay, decaying at a rate of 0.9 at the end of every epoch. Compared with other schedules, exponential learning rate trains the model more gradually, which helps to tune when the teacher should be early-stopped. For the student, we use the OneCycle learning rate~\citesupp{smith2019super} with two phases and 0.21 as the maximum learning rate, updated every batch. To train the teacher and the student, we use batch size 128 and 256, respectively.

All adversarial examples are crafted with L$_{\infty}$ perturbations of size $\varepsilon=8/255$. For all models trained with adversarial training or AKD, we craft the adversarial examples using PGD-7, unless stated otherwise. To evaluate robustness, we use PGD-7~\cite{madry2017towards} and AutoAttack~\cite{croce2020reliable}.

\subsection{Generalizing to RSLAD}

The original loss function of RSLAD~\cite{zi2021revisiting} is
\begin{equation}
    \text{RSLAD}(\bm x) = (1 - \lambda)\text{KL}(f_S(\bm x), f_T(\bm x)) + \lambda\text{KL}(f_S(\bm x'), f_T(\bm x)).
    \label{eq:rslad}
\end{equation}

To incorporate label mixing into RSLAD (see Tab.~\ref{tab:other_distill}), we have used the following formulation:
\begin{align*}
    \text{RSLAD}(\bm x) = (1 - \lambda)&\text{KL}(f_S(\bm x), \alpha f_T(\bm x) + (1-\alpha)y) \\
    + \; \lambda&\text{KL}(f_S(\bm x'), \alpha f_T(\bm x) + (1-\alpha)y)
    \label{eq:rslad}
\end{align*}

\section{Early stopping and label mixing}

\subsection{Trade-offs for RSLAD}


We have illustrated that early stopping the teacher and label mixing is beneficial for AKD in Sections~\ref{sec:es} and \ref{sec:alpha}. To see if this is not unique to our method, and can be applied to other robust distillation methods, we make the same analysis for RSLAD. This analysis is complementary with the results on Table~\ref{tab:other_distill}.

\begin{figure}[t]
    \centering
    \subfloat[Teacher standardly trained]{
        \centering
        \includegraphics[width=0.45\textwidth]{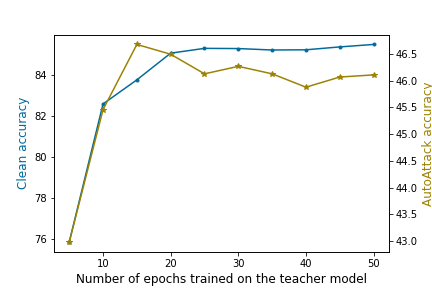}
    }
    \subfloat[Teacher adversarially trained]{
        \centering
        \includegraphics[width=0.45\textwidth]{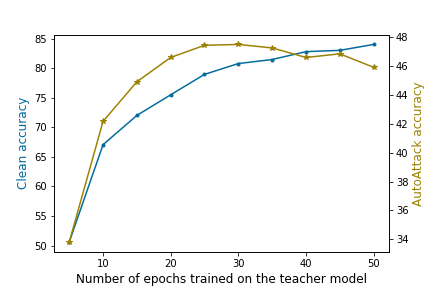}
    }
    \caption{Effect of early stopping the teacher on the RSLAD method. We used a standardly (left) and an adversarially (right) trained ResNet-18 for the teacher model, and a ResNet-18 for the student, on the CIFAR10 dataset.}
    \label{fig:rslad_es}
\end{figure}

We compare in Figure~\ref{fig:rslad_es} how the performance of the student model varies depending on when we early-stopped its teacher. Here, the student is trained with AKD using $\alpha=1.0$ to avoid any confounding effect from label mixing. We see that in both cases, we find an optimal point where robustness is maximized, consistent with the results in Figure~\ref{fig:akd_es}. Moreover, we observe that the optimal early stopping point is found earlier when the teacher has been standardly trained, probably due to its faster convergence speed.

\begin{figure}[t]
    \centering
    \subfloat[Teacher standardly trained]{
        \centering
        \includegraphics[width=0.45\textwidth]{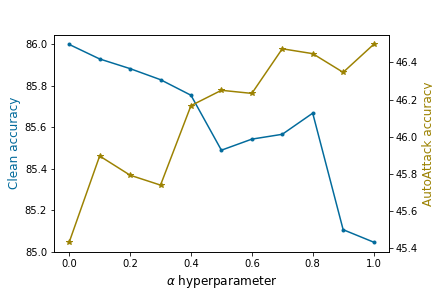}
    }
    \subfloat[Teacher adversarially trained]{
        \centering
        \includegraphics[width=0.45\textwidth]{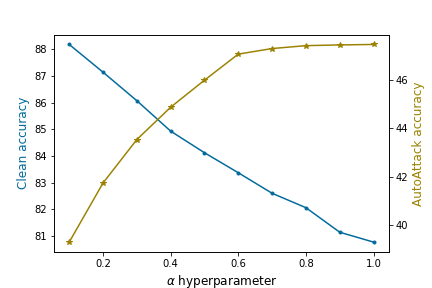}
    }
    \caption{Effect of label mixing on the RSLAD method. We used a standardly (left) and an adversarially (right) trained ResNet-18 for the teacher model, and a ResNet-18 for the student, on the CIFAR10 dataset.}
    \label{fig:rslad_alpha}
\end{figure}

We compare in Figure~\ref{fig:rslad_alpha} the performance of the student model when we change the value of $\alpha$. We consider a standardly trained teacher and an adversarially trained teacher, early-stopped at epochs 20 and 30, respectively. We highlight that, unlike AKD, RSLAD uses the prediction of the teacher evaluated on the clean sample. Thus, for standardly trained teachers, even when using mostly the distilled labels, we do not lose the robustness completely, which contrasts with the results in Figure~\ref{fig:akd_alpha}. Furthermore, both standardly and adversarially trained teachers create a similar behavior: using the original labels increases the accuracy of the student, while using the distilled labels increases robustness instead. Thus, the better trade-offs would use values of alpha close to 0.5, where we have a better trade-off between robustness and accuracy.

\subsection{Trade-offs for compression}

We have illustrated that early stopping the teacher and label mixing is beneficial when used to boost the performance of similar capacity models in Sections~\ref{sec:es} and \ref{sec:alpha}. We show that these insights also generalize in the context of compression, where the teacher model is significantly bigger than the student model.

\begin{figure}[t]
    \centering
    \subfloat[Early stopping effect]{
        \centering
        \includegraphics[width=0.45\textwidth]{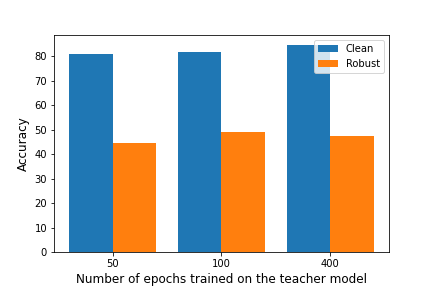}
        \label{fig:sota_es}
    }
    \subfloat[Label mixing effect]{
        \centering
        \includegraphics[width=0.45\textwidth]{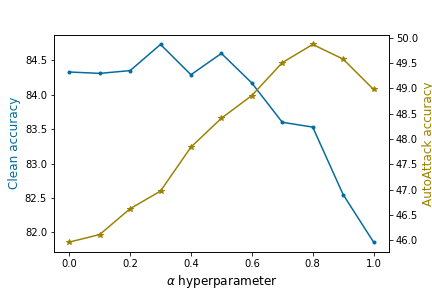}
        \label{fig:sota_alpha}
    }
    \caption{Effect of early stopping the teacher (left) and label mixing (right) on the AKD method. We use the state-of-the-art WideResNet-28-10 with swish activations from \cite{rebuffi2021fixing} as the teacher, and the ResNet-18 for the student, on the CIFAR10 dataset.}
    \label{fig:sota_trade-offs}
\end{figure}

We compare in Figure \ref{fig:sota_es} how the performance of the student model varies depending on when we early-stopped its teacher. For the student, we use the ResNet-18 architecture, while for the teacher we use the WideResNet-28~\cite{he2016deep,zagoruyko2016wide} with swish activations~\cite{hendrycks2016gaussian} architecture, as proposed in~\cite{rebuffi2021fixing}. The student is trained with AKD, using $\alpha=1.0$ to avoid any confounding effect from label mixing. Meanwhile, the teacher is trained using model weight averaging~\cite{izmailov2018averaging}, extra data sampled from generative models~\cite{ho2020denoising} and the default data augmentation techniques~\cite{devries2017improved} given in~\cite{rebuffi2021fixing}. From this plot, we can observe there is an optimal early stopping point where the student robustness is maximized. In contrast, we also see that distilling from an early-stopped teacher reduces the student clean accuracy. These results are consistent with the ones we obtained when using a smaller, adversarially trained, ResNet-18 teacher (see Figure~\ref{fig:akd_es}).

We compare in Figure~\ref{fig:akd_alpha} the performance of the student model when we change the value of $\alpha$. We consider the teacher has been early-stopped at epoch 100. We observe that the student robustness is maximized with a high $\alpha$ value. This result is consistent with Figure~\ref{fig:akd_alpha}, in which the teacher has the same capacity as the student. However, in contrast, we find that the student network robustness is much higher, thus showing that AKD can be used for compressing the superior performance of a bigger network, like other RKD methods.

\section{Teacher robustness trade-offs for AKD}

\subsection{Label mixing and early stopping conservative trade-offs}

For the results we have presented for differently trained teachers in Table~\ref{tab:akd_teacher_training}, we optimize the early stopping epoch of the teacher and the label mixing parameter, to maximize robust performance when the teacher is trained adversarially, and clean performance otherwise. Here, we present the results we obtained when the objective is not to maximize one metric, but to maximize it while reducing the cost to the other. 

We show in Tables~\ref{tab:akd_typeof_source_extra} and \ref{tab:akd_typeof_source_cifar100} the comparison between using early-stopping and label mixing parameters that give more conservative trade-offs (top rows) and using parameters that maximize one metric (bottom rows). We show these results for different trained teachers and between using one model and an ensemble on the CIFAR10 and CIFAR100 datasets. We find that using $\alpha=0.5$ and early-stopping the teacher at the end of the 20th epoch, gives better clean-robust accuracy trade-offs at the cost of not maximizing one particular metric. Particularly, for standardly trained models, using the slightly smaller $\alpha$ value can mitigate greatly the reduction of robust accuracy, that happens when using high values of $\alpha$ when distilling non-robust models using AKD. For adversarially trained models, there is less benefit to using a conservative trade-off, which increments accuracy at a similar cost to robustness.

We compare in Tables~\ref{tab:small_eps_cifar10} and \ref{tab:small_eps_cifar100} how the performance of the student varies depending on the $\varepsilon$ size of the attack used to train the teacher model adversarially for the CIFAR10 and CIFAR100 datasets. We show that, for a small $\varepsilon$, we can get better accuracy and robustness than when we used standard training. This increase in generalization can be transferred to the student, even when the student is still trained using perturbations with the standard $\varepsilon$ value of 8/255 used in CIFAR datasets.

\begin{table*}[t]
    \centering
    \caption{Effect of using different early-stopping and label mixing configurations for the teacher model on the AKD method. At the top, we provide the more conservative configurations, while at the bottom we show the ones that maximize clean or robust performance. We use the ResNet-18 architecture for both student and teacher models on the CIFAR10 dataset. For all adversarial attacks, we use $L_{\infty}$ norm of size 8/255. The results show the average of four different runs, with one standard deviation in smaller font.}
    \label{tab:akd_typeof_source_extra}
    \begin{tabular}{ccc|cc|cc}
    \toprule
        \multicolumn{3}{c|}{} & \multicolumn{2}{c}{1 model} & \multicolumn{2}{c}{4 models}\\
    	    \cline{4-7}
          Teacher training & ES & $\alpha$ & Clean & Robust & Clean & Robust \\
         \hline
         Std. & ep. 20 & 0.5 & $86.28_{\pm0.04}$ & $44.63_{\pm0.61}$ & $86.91_{\pm0.15}$ & $45.64_{\pm0.20}$ \\
         AT$_{\text{FFGSM}}$ & ep. 20 & 0.5 & $83.80_{\pm0.30}$ & $47.05_{\pm0.15}$ & $84.05_{\pm0.10}$ & $47.51_{\pm0.14}$ \\
         AT$_{\text{PGD-7}}$ & ep. 20 & 0.5 & $83.13_{\pm0.26}$ & $47.33_{\pm0.13}$ & $83.48_{\pm0.13}$ & $47.61_{\pm0.13}$ \\
         \midrule
         Std. & ep. 20 & 0.6 & $86.22_{\pm0.43}$ & $39.80_{\pm1.53}$ & $87.27_{\pm0.12}$ & $43.36_{\pm0.13}$ \\
         AT$_{\text{FFGSM}}$ & ep. 30 & 0.75 & $83.52_{\pm0.41}$ & $46.99_{\pm0.17}$ & $83.90_{\pm0.05}$ & $47.47_{\pm0.17}$ \\
         AT$_{\text{PGD-7}}$ & ep. 30 & 0.75 & $82.70_{\pm0.15}$ & $47.66_{\pm0.17}$ & $82.96_{\pm0.12}$ & $47.96_{\pm0.15}$ \\
         \bottomrule
    \end{tabular}
\end{table*}

\begin{table*}[t]
    \centering
    \caption{Effect of using different early-stopping and label mixing configurations for the teacher model on the AKD method. At the top, we provide the more conservative configurations, while at the bottom we show the ones that maximize clean or robust performance. We use the ResNet-18 architecture for both student and teacher models on the CIFAR100 dataset. For all adversarial attacks, we use $L_{\infty}$ norm of size 8/255.}
    \label{tab:akd_typeof_source_cifar100}
    \begin{tabular}{ccc|cc|cc}
    \toprule
        \multicolumn{3}{c|}{} & \multicolumn{2}{c}{1 model} & \multicolumn{2}{c}{4 models}\\
    	    \cline{4-7}
          Teacher training & ES & $\alpha$ & Clean & AutoAttack & Clean & AutoAttack \\
         \hline
         Std. & ep. 20 & 0.5 & $61.90$ & $21.23$ & $63.11$ & $21.98$ \\
         AT$_{\text{FFGSM}}$ & ep. 20 & 0.5 & $58.38$ & $23.44$ & $58.79$ & $24.64$ \\
         AT$_{\text{PGD-7}}$ & ep. 20 & 0.5 & $57.74$ & $23.59$ & $58.02$ & $24.27$ \\
         \midrule
         Std. & ep. 20 & 0.6 & $62.09$ & $19.56$ & $63.72$ & $20.85$ \\
         AT$_{\text{FFGSM}}$ & ep. 30 & 0.75 & $60.51$ & $20.93$ & $62.69$ & $22.26$ \\
         AT$_{\text{PGD-7}}$ & ep. 30 & 0.75 & $56.87$ & $24.61$ & $58.34$ & $24.93$ \\
         \bottomrule
    \end{tabular}
\end{table*}

\begin{table}[t]
    \centering
    \caption{Effect of using different $L_\infty$ ball constraints to adversarially train the teacher model on the AKD method. At the top, we provide the more conservative early-stopping and label mixing configurations, while at the bottom we show the ones that maximize clean or robust performance. We use the ResNet-18 architecture for both teacher and student models on the CIFAR10 dataset.}
    \label{tab:small_eps_cifar10}
    \begin{tabular}{ccc|cc}
    \toprule
          $\alpha$ & ES & $L_\infty$ ball size ($\varepsilon$) & Clean & AutoAttack \\
         \hline
         $0.5$ & ep. 20 & 0 & $86.28_{\pm0.04}$ & $44.63_{\pm0.61}$ \\
         $0.5$ & ep. 20 & 2/255 & $86.25_{\pm0.31}$ & $46.13_{\pm0.43}$ \\
         $0.5$ & ep. 20 & 8/255 & $83.13_{\pm0.26}$ & $47.33_{\pm0.13}$ \\
         $0.5$ & ep. 20 & 16/255 & $82.15_{\pm0.23}$ & $47.25_{\pm0.11}$ \\
         \midrule
         $0.6$ & ep. 20 & 0 & $86.22_{\pm0.43}$ & $39.80_{\pm1.53}$ \\
         $0.6$ & ep. 20 & 2/255 & $86.59_{\pm0.22}$ & $45.42_{\pm0.18}$ \\
         $0.75$ & ep. 30 & 8/255 & $82.70_{\pm0.15}$ & $47.66_{\pm0.17}$ \\
         $0.75$ & ep. 30 & 16/255 & $76.76_{\pm0.33}$ & $47.43_{\pm0.16}$ \\
         \bottomrule
    \end{tabular}
\end{table}

\begin{table}[t]
    \centering
    \caption{Effect of using different $L_\infty$ ball constraints to adversarially train the teacher model on the AKD method. At the top, we provide the more conservative early-stopping and label mixing configurations, while at the bottom we show the ones that maximize clean or robust performance. We use the ResNet-18 architecture for both teacher and student models on the CIFAR100 dataset.}
    \label{tab:small_eps_cifar100}
    \begin{tabular}{ccc|cc}
    \toprule
          $\alpha$ & ES & $L_\infty$ ball size ($\varepsilon$) & Clean & AutoAttack \\
         \hline
         $0.5$ & ep. 20 & 0 & $61.90$ & $21.23$ \\
         $0.5$ & ep. 20 & 2/255 & $61.64$ & $22.33$ \\
         $0.5$ & ep. 20 & 8/255 & $57.74$ & $23.59$ \\
         \midrule
         $0.6$ & ep. 20 & 0 & $62.09$ & $19.56$ \\
         $0.6$ & ep. 20 & 2/255 & $62.09$ & $22.12$ \\
         $0.75$ & ep. 30 & 8/255 & $56.87$ & $24.61$ \\
         \bottomrule
    \end{tabular}
\end{table}

\subsection{Ensemble of teachers performance}

We study the effect of using an ensemble of teachers in Section~\ref{sec:ensemble}, and show that they help improve the student performance. Here, we show that this increase in performance is not caused by the teacher itself being more robust or accurate. In fact, in Table~\ref{tab:ensemble} we show that using an ensemble of adversarially trained models does not improve the robustness or the clean accuracy compared to using one model. Strangely, when using AutoAttack to evaluate the ensemble, we find that the AutoAttack's black-box attacks were ineffective. The ensemble reports an AutoAttack accuracy of 51.58\% compared to the 46.99\% obtained with a single model. Due to the small difference between the reported AutoAttack and PGD-7 accuracies of the ensemble, we believe further evaluation is required using adaptive black-box attacks that take the ensemble into account \citesupp{he2017adversarial}.

\begin{table}[t]
    \caption{Comparison between the performance of one adversarially trained ResNet-18 model and an ensemble of four adversarially trained ResNet-18 models using different inititializations.}
    \label{tab:ensemble}
    \centering
    \begin{tabular}{c|cc}
         \toprule
          Model & Clean & PGD-7 \\
         \hline
         ResNet-18 (AT) & 84.22 & 53.56 \\
         Ensemble ResNet-18 (AT) & 81.86 & 52.74 \\
         \bottomrule
    \end{tabular}
\end{table}

\section{Sample difficulty}

\begin{figure}[t]
    \centering
    \subfloat[Easiest to learn]{
        \centering
        \includegraphics[width=1.0\textwidth]{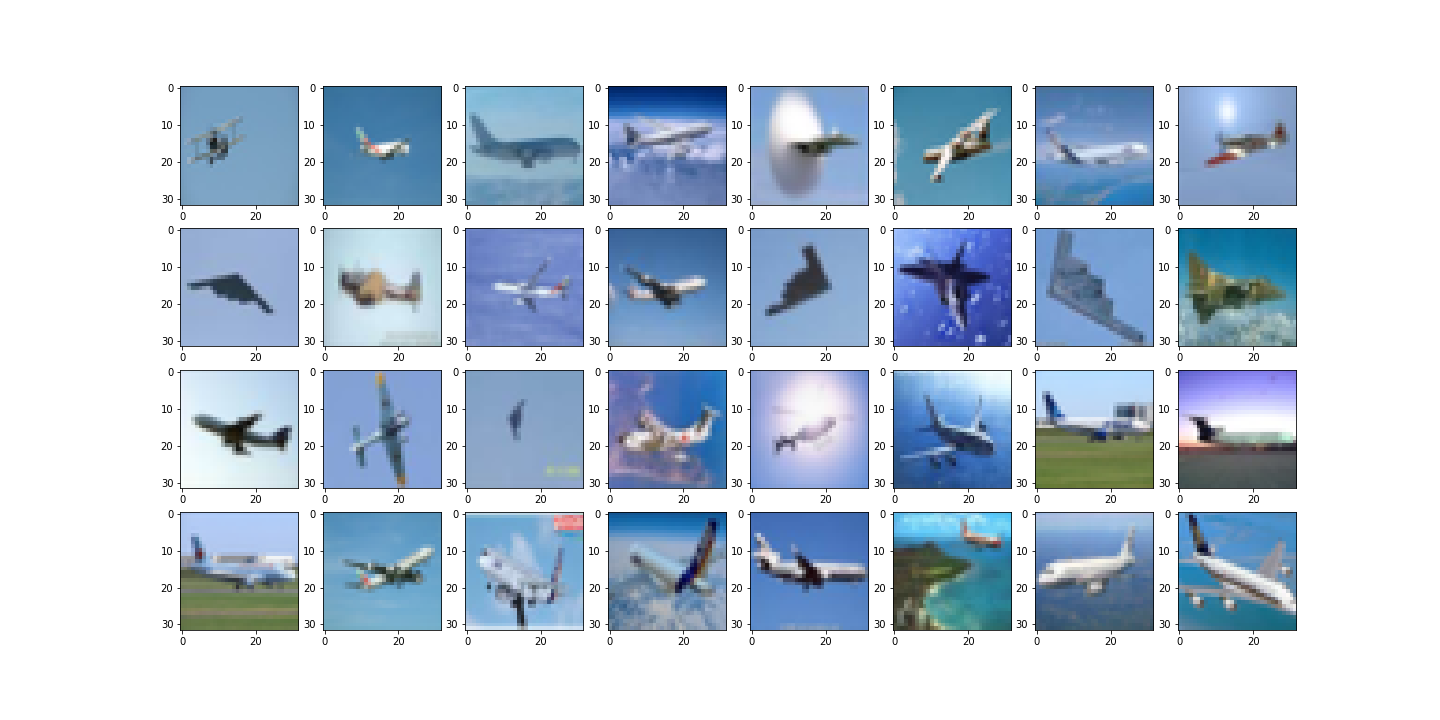}
    } \\
    \subfloat[Hardest to learn]{
        \centering
        \includegraphics[width=1.0\textwidth]{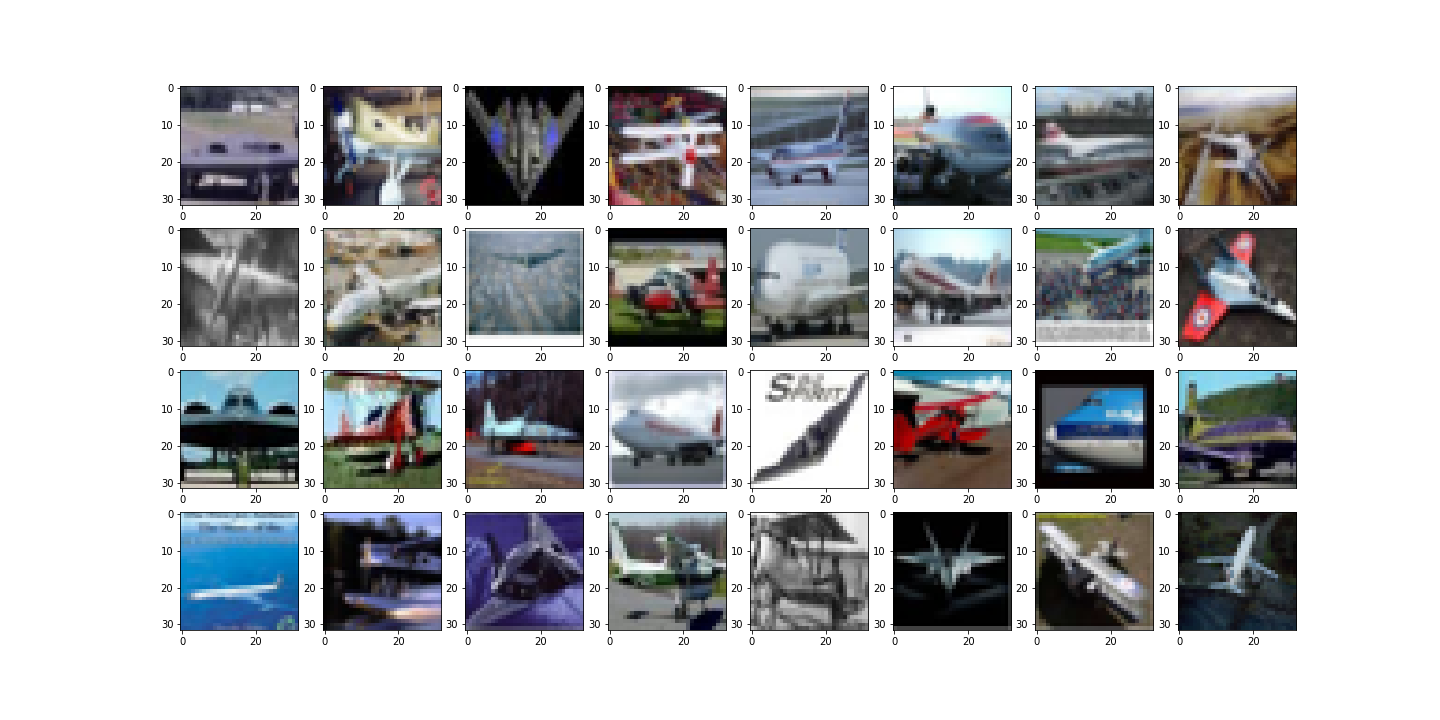}
    }
    \caption{Easiest and hardest to learn plane images based on the difficulty metric from~\cite{liu2021impact,dong2021data} applied to a standardly trained model on a subset of CIFAR10 containing only planes and cars.}
    \label{fig:airplanes}
\end{figure}

\begin{figure}[t]
    \centering
    \subfloat[Easiest to learn]{
        \centering
        \includegraphics[width=1.0\textwidth]{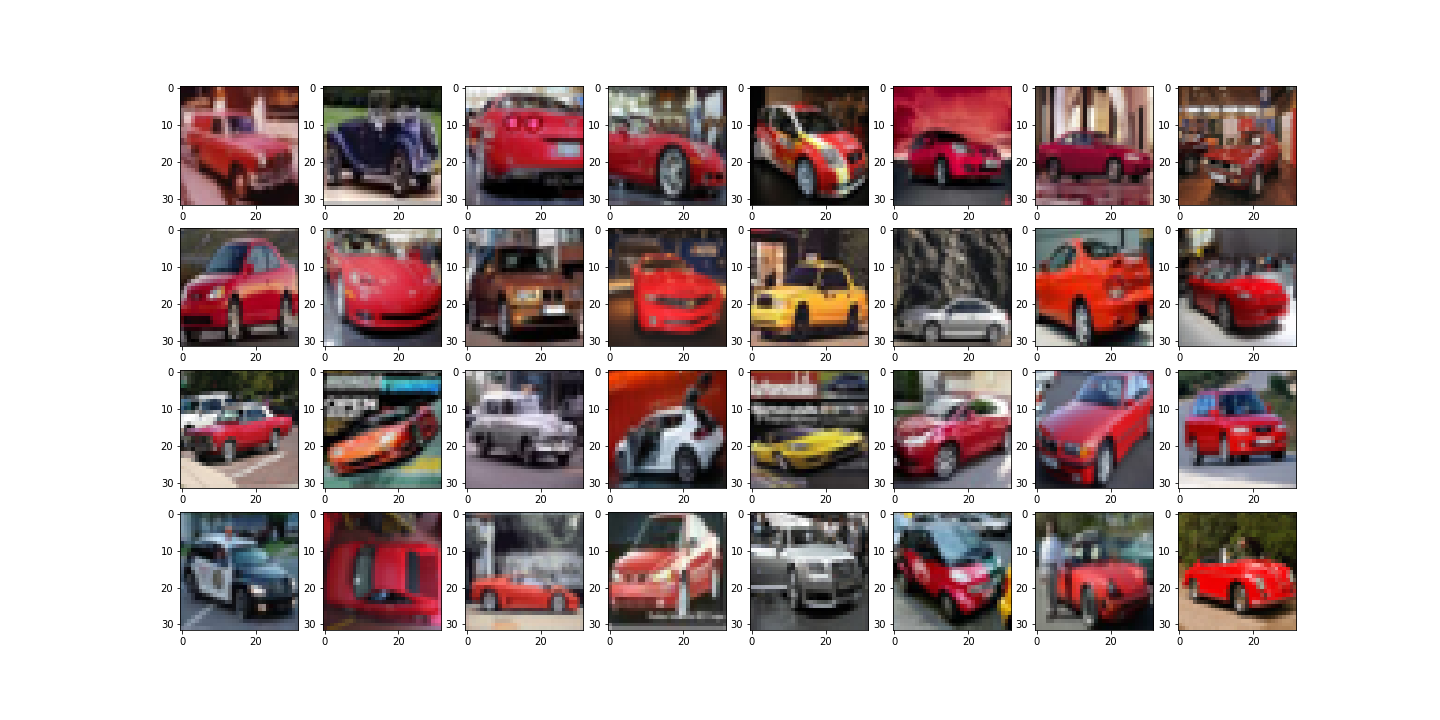}
    } \\
    \subfloat[Hardest to learn]{
        \centering
        \includegraphics[width=1.0\textwidth]{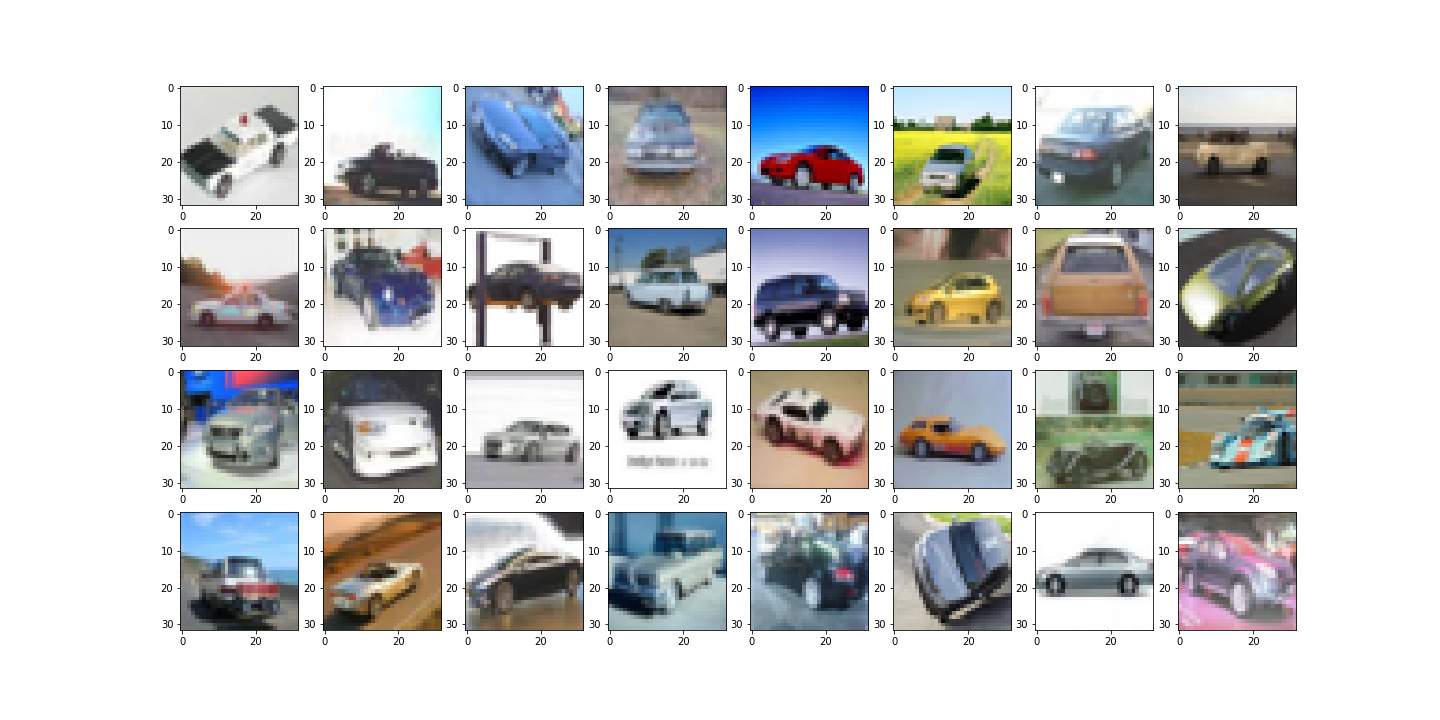}
    }
    \caption{Easiest and hardest to learn car images based on the difficulty metric from~\cite{liu2021impact,dong2021data} applied to a standardly trained model on a subset of CIFAR10 containing only planes and cars.}
    \label{fig:cars}
\end{figure}

In this Section we show some qualitative results that validate following the practice in~\cite{liu2021impact,dong2021data} to rank how difficult each sample is to learn. We apply the sample difficulty formulation \eqref{eq:difficulty} to a model standardly trained on the experimental settings we gave in Section~\ref{sec:trajectories}. We plot the 32 ``easiest" and ``hardest" images of planes and cars in Figures~\ref{fig:airplanes} and \ref{fig:cars}, respectively. Qualitatively, we find that the network distinguishes between car and plane images based on the overall color of the image and how busy is the background. An image with a lot of blue and light colors and simple sky backgrounds most probably is classified as a plane, while complex backgrounds with mostly red and dark colors are classified as car images. Thus, the network in general finds more difficulty on images that present uncommon features or features from the other class. This is shown on our selected difficult images: planes with complicated backgrounds and cars with sky-like backgrounds.

\section{Training trajectories}

We will now show extended results on the impact of AKD on the final performance and training trajectories of samples of different difficulty. The experimental settings are the same as the ones described in Section~\ref{sec:trajectories}.

\subsection{Adversarially trained teacher}

\begin{figure}[t]
    \centering
    \subfloat[$p_{T,K}$]{
        \centering
        \includegraphics[width=0.45\textwidth]{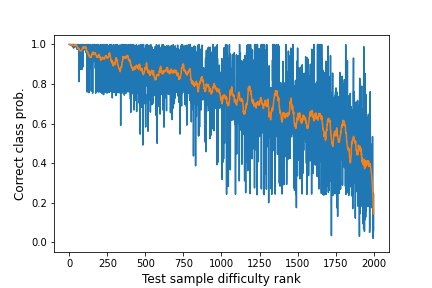}
    }
    \subfloat[$p_{S,K} - p_{T,K}$]{
        \centering
        \includegraphics[width=0.45\textwidth]{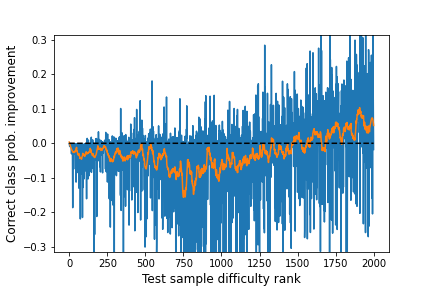}
    }
    \caption{Correct class probability of the adversarially trained teacher (left) and the relative improvement of the student (right) tested on natural images. The blue line shows the true values, and the orange line a smoothed-out version.}
    \label{fig:all_at_akd_distil_std}
\end{figure}

We show in Section~\ref{sec:trajectories} the calibration effect of AKD on the student model, increasing performance on hard samples at the cost of confidence on easier ones. Here, we show in Figure \ref{fig:all_at_akd_distil_std} that the effect AKD has on clean performance is quite similar. However, when compared with the results in terms of robustness we show in Figure~\ref{fig:all_at_akd_distil}, it is clear that AKD is more beneficial in terms of robustness, where more difficult samples improve and the improvement is more significant.

\begin{figure}[t]
    \centering
    \subfloat[Nat. images]{
        \centering
        \includegraphics[width=0.45\textwidth]{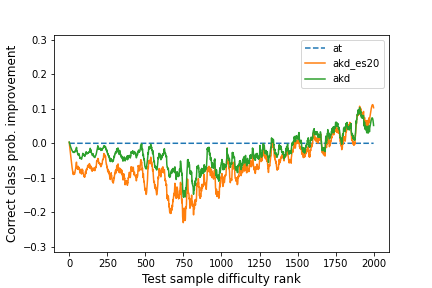}
        \label{fig:es_all_at_imp}
    }\quad
    \subfloat[Adv. images]{
        \centering
        \includegraphics[width=0.45\textwidth]{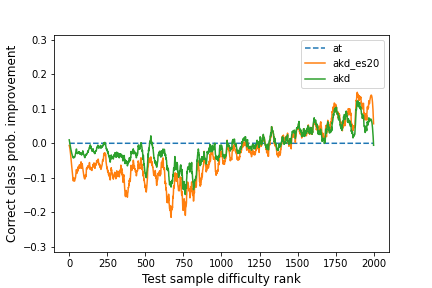}
        \label{fig:es_all_at_imp_at}
    }
    \caption{Smoothed improvement of the correct class probability ($p_{S,K} - p_{T,K}$) when using AKD with and without early stopping an adversarially trained teacher. We evaluate the performance on natural images (left) and adversarial ones (right).}
    \label{fig:es_all_at}
\end{figure}

\begin{figure}[t]
    \centering
    \subfloat[Nat. images]{
        \centering
        \includegraphics[width=0.45\textwidth]{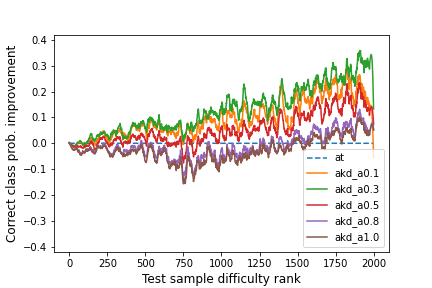}
    } \quad
    \subfloat[Adv. images]{
        \centering
        \includegraphics[width=0.45\textwidth]{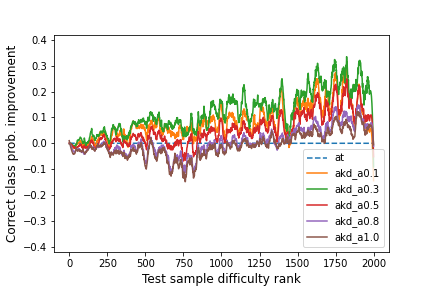}
    }
    \caption{Smoothed improvement of the correct class probability ($p_{S,K} - p_{T,K}$) when using AKD with an adversarially trained teacher, for different values of the label mixing parameter $\alpha$. We evaluate the performance on natural images (left) and adversarial ones (right).}
    \label{fig:alpha_all_at}
\end{figure}

We also replicate this analysis using early-stopping and label mixing. We show in Figure \ref{fig:es_all_at_imp} that early-stopping causes the accuracy to only increase for the most difficult samples, while reducing performance for most of them. As shown in Figure \ref{fig:es_all_at_imp_at}, for robustness it is a bit less detrimental, but the effect is quite similar.
We show in Figure \ref{fig:alpha_all_at} the same analysis, but for different values of the label mixing parameter $\alpha$. We find that label mixing increases clean and robust performance for all samples, but the effect increases proportionally with the sample difficulty. In this particular setting, using $\alpha=0.3$ results in a optimum increase in performance.

\subsection{Standardly trained teacher}

\begin{figure}[t]
    \centering
    \subfloat[Nat. images]{
        \centering
        \includegraphics[width=0.45\textwidth]{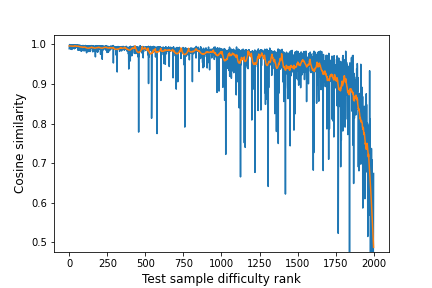}
    }
    \subfloat[Adv. images]{
        \centering
        \includegraphics[width=0.45\textwidth]{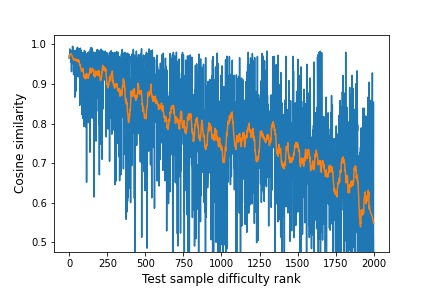}
    }
    \caption{Cosine similarity when using AKD with an standardly trained teacher when evaluated on natural (left) and adversarial (right) images. The blue line shows the true values, and the orange line a smoothed-out version.}
    \label{fig:trj_all_cossim}
\end{figure}

We show the difference between training dynamics between the teacher and the student, when using AKD with a standardly trained teacher. Figure \ref{fig:trj_all_cossim} shows precisely these differences on all natural and adversarial test samples, ordered in terms of learning difficulty. 

\begin{figure}[t]
    \centering
    \subfloat[$p_{T,K}$]{
        \centering
        \includegraphics[width=0.45\textwidth]{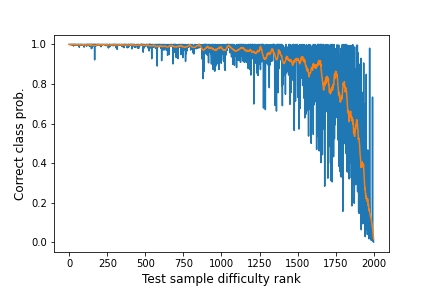}
    }
    \subfloat[$p_{S,K} - p_{T,K}$]{
        \centering
        \includegraphics[width=0.45\textwidth]{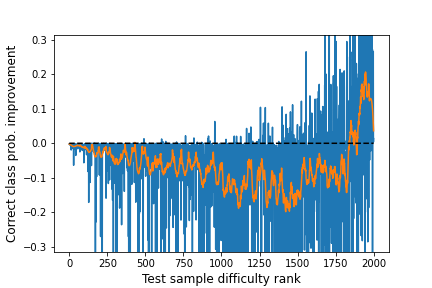}
    }
    \caption{Correct class probability of the standardly trained teacher (left) and the relative improvement of the student (right) tested on natural images. The blue line shows the true values, and the orange line a smoothed-out version.}
    \label{fig:all_akd_distil}
\end{figure}

Compared with Figure \ref{fig:all_at_akd_cossim}, we see a similar behavior for the natural images. That is, the easier-to-learn samples mostly follow the same trajectory on the student and teacher models, while the training trajectories clearly differ on hard examples. When looking at the improvement on Figure~\ref{fig:all_akd_distil}, we see a large improvement for hard samples, at the cost of confidence on easier ones. Compared with Figure~\ref{fig:all_at_akd_distil_std}, the improvement is large thanks to the teacher being standardly trained, which transfer clean accuracy better than an adversarially trained teacher.

\begin{figure}[t]
    \centering
    \subfloat[$p_{T,K}$]{
        \centering
        \includegraphics[width=0.45\textwidth]{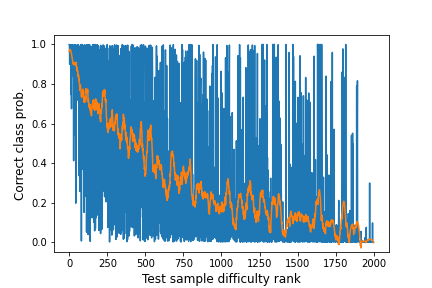}
    }
    \subfloat[$p_{S,K} - p_{T,K}$]{
        \centering
        \includegraphics[width=0.45\textwidth]{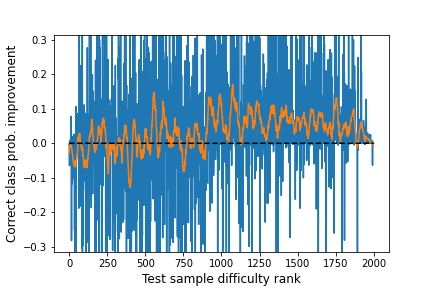}
    }
    \caption{Correct class probability of the standardly trained teacher (left) and the relative improvement of the student (right) tested on adversarial images. The blue line shows the true values, and the orange line a smoothed-out version.}
    \label{fig:all_akd_distil_at}
\end{figure}

However, in contrast, the adversarial images training trajectories change a lot in general, especially for hard samples (see Figure~\ref{fig:trj_all_cossim}). This can be explained by the settings of the experiments. Since we are evaluating the effect of AKD in itself, we do not use any label mixing. We find that the difference in trajectories is slightly correlated with the teacher performance. Thus, the trajectories will be more similar for the easier samples, which the teacher classifies correctly due to the simplicity of the task (see Figure \ref{fig:all_akd_distil_at}). But as the samples are harder, the robustness of the teacher drops, with makes the trajectories also more dissimilar. We can see in Figure \ref{fig:all_akd_distil_at} that AKD helps increase slightly the confidence for all samples, but it does not correct misclassification by itself.

\begin{figure}[t]
    \centering
    \subfloat[Nat. images]{
        \centering
        \includegraphics[width=0.45\textwidth]{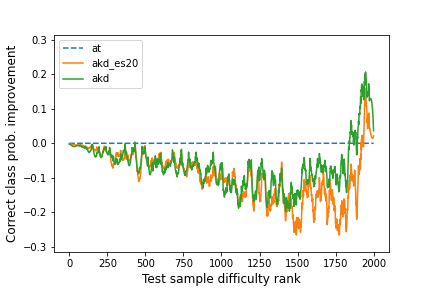}
        \label{fig:es_all_imp}
    }\quad
    \subfloat[Adv. images]{
        \centering
        \includegraphics[width=0.45\textwidth]{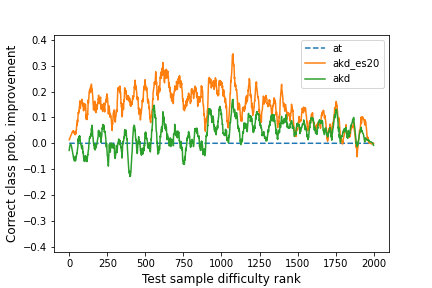}
        \label{fig:es_all_imp_at}
    }
    \caption{Smoothed improvement of the correct class probability ($p_{S,K} - p_{T,K}$) when using AKD with and without early stopping a standardly trained teacher. We evaluate the performance on natural images (left) and adversarial ones (right).}
    \label{fig:es_all}
\end{figure}

We also replicate this analysis using early-stopping and label mixing. We show in Figure \ref{fig:es_all_imp} that early-stopping reduces the accuracy for hard samples. This reduction is consistent with~\cite{cho2019efficacy}, where they claim that for CKD, early stopping does not help improve clean performance when there is no capacity mismatch between the teacher and student models. In contrast, robustness increases for easier samples, as shown in Figure \ref{fig:es_all_at_imp_at}.

We show in Figure \ref{fig:alpha_all_at} the same analysis, but for different values of the label mixing parameter $\alpha$. We find that label mixing increases clean performance for all samples, but the effect increases proportionally with the sample difficulty. In terms of robust performance, we see the same behavior we presented in Figure~\ref{fig:akd_alpha} when using a standardly trained model. That is, for high values of $\alpha$, the student is not robust, while for small values the robustness is high. This analysis shows it is a bit more nuanced than the previously presented results, and for high values of $\alpha$ the model improves a bit its confidence in the correct class, but not enough to avoid misclassification.

\begin{figure}[t]
    \centering
    \subfloat[Nat. images]{
        \centering
        \includegraphics[width=0.45\textwidth]{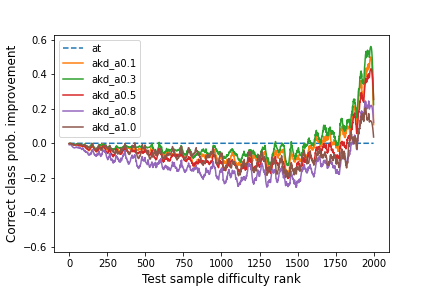}
    } \quad
    \subfloat[Adv. images]{
        \centering
        \includegraphics[width=0.45\textwidth]{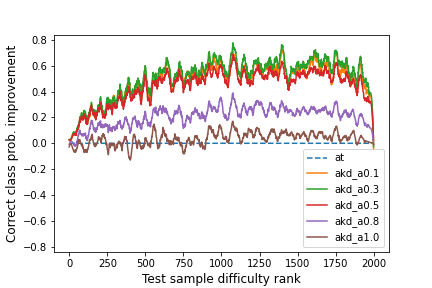}
    }
    \caption{Smoothed improvement of the correct class probability ($p_{S,K} - p_{T,K}$) when using AKD with a standardly trained teacher, for different values of the label mixing parameter $\alpha$. We evaluate the performance on natural images (left) and adversarial ones (right).}
    \label{fig:alpha_all}
\end{figure}